\documentclass{article}

\usepackage{tabularx}

\usepackage{arxiv}

\usepackage[utf8]{inputenc} 
\usepackage[T1]{fontenc}    
\usepackage{hyperref}       
\usepackage{url}            
\usepackage{booktabs}       
\usepackage{amsfonts}       
\usepackage{nicefrac}       
\usepackage{microtype}      
\usepackage{lipsum}		
\usepackage{graphicx}
\usepackage{natbib}
\usepackage{doi}

\usepackage{algorithm}
\usepackage{algpseudocode}
\usepackage{mathtools}

\usepackage{multirow}
\usepackage{booktabs}
\usepackage{xspace}
\usepackage{fancyhdr}

\newcommand{\ours}{JL-GAT\xspace}

\title{Joint-Local Grounded Action Transformation for Sim-to-Real Transfer in Multi-Agent Traffic Control}

\author{
  Justin Turnau \\
  \texttt{jturnau@asu.edu} \\
  \And
  Longchao Da \\
  \texttt{longchao@asu.edu} \\
  \And
  Khoa Vo \\
  \texttt{ngocbach@asu.edu} \\
  \And
  Ferdous Al Rafi\textsuperscript{2} \\
  \texttt{rafirafi155@gmail.com} \\
  \And
  Shreyas Bachiraju \\
  \texttt{sbachira@asu.edu} \\
  \And
  Tiejin Chen \\
  \texttt{tchen169@asu.edu} \\
  \And
  Hua Wei \\
  \texttt{hua.wei@asu.edu} \\
}

\renewcommand{\shorttitle}{Joint-Local Grounded Action Transformation}

\date{}
\begin{document}
\maketitle
\begin{abstract}
Traffic Signal Control (TSC) is essential for managing urban traffic flow and reducing congestion. Reinforcement Learning (RL) offers an adaptive method for TSC by responding to dynamic traffic patterns, with multi-agent RL (MARL) gaining traction as intersections naturally function as coordinated agents. However, due to shifts in environmental dynamics, implementing MARL-based TSC policies in the real world often leads to a significant performance drop, known as the sim-to-real gap. Grounded Action Transformation (GAT) has successfully mitigated this gap in single-agent RL for TSC, but real-world traffic networks, which involve numerous interacting intersections, are better suited to a MARL framework. In this work, we introduce \ours, an application of GAT to MARL-based TSC that balances scalability with enhanced grounding capability by incorporating information from neighboring agents. \ours adopts a decentralized approach to GAT, allowing for the scalability often required in real-world traffic networks while still capturing key interactions between agents. Comprehensive experiments on various road networks under simulated adverse weather conditions, along with ablation studies, demonstrate the effectiveness of \ours. The code is publicly available at \href{https://github.com/DaRL-LibSignal/JL-GAT/}{https://github.com/DaRL-LibSignal/JL-GAT/}.
\end{abstract}

\footnotetext[1]{Arizona State University, Tempe, AZ, USA}
\footnotetext[2]{Bangladesh University of Engineering and Technology, Dhaka, Bangladesh}


\section{Introduction}
\label{sec:introduction}

Reinforcement Learning (RL) is well-suited for sequential decision-making, enabling agents to learn effective policies through interaction with the environment~\citep{roijers2013survey}. This data-driven design, together with the ability to adaptively refine policies, makes RL a powerful approach to complex real-world problems. Traffic Signal Control (TSC) is an effective way to reduce congestion, minimize travel times, and improve urban mobility~\citep{wei2018intellilight}. By modeling TSC as a sequential decision-making problem, where each traffic signal chooses timing and phases based on evolving traffic conditions, RL can deliver flexible, efficient control strategies. Thus, RL-driven TSC appears as a dynamic and robust alternative to static or rule-based methods in transportation research~\citep{wei2019survey}.

In addition to treating an intersection-coupled traffic signal as a single agent, multi-agent reinforcement learning (MARL) is essential for scaling up traffic signal control to complex urban networks~\citep{jiang2024x}. By deploying a network of agents, each controlling an individual intersection, MARL facilitates decentralized decision-making while maintaining coordinated across the entire system~\citep{chen2020toward}. It allows each agent to learn local policies that are responsive to immediate traffic conditions yet also adapt through communication and cooperation with neighboring agents to optimize overall traffic flow, which is more suitable for managing large-scale, dynamic transportation environments such as those found in real-world applications~\citep{balmer2004large}.

In order to learn the traffic signal control policies, a direct way is to leverage the existing traffic simulators (e.g., SUMO~\citep{behrisch2011sumo}, CityFlow~\citep{zhang2019cityflow, da2024cityflower}) as an interactive environment and explore control policies. While simulators offer a controlled environment to train and evaluate RL-based TSC policies, transitioning these models from simulation to the real world introduces a challenging gap known as the sim-to-real issue~\citep{da2023sim2real}. Discrepancies between the simulated and real environments, such as unmodeled traffic dynamics~\citep{da2023uncertainty}, sensor noise~\citep{qadri2020state}, and unpredictable driver behaviors~\citep{lee2015necessity}, can lead to significant deviations in performance. Therefore, robust sim-to-real techniques are essential to bridge this gap and ensure the performance observed in simulation translates to real-world urban settings.

The preliminary research from~\citep{da2023sim2real} has identified the severity of the sim-to-real issue in RL-based TSC. There are several proposed solutions to mitigate the sim-to-real gap, either by calibrating the simulator's realism~\citep{muller2021towards} or by using transfer learning in the RL training paradigm, such as grounded action transformation (GAT)~\citep{da2024prompt}.
\ours enhances GAT by integrating neighboring agents' information to capture local interactions, improving transition dynamics modeling. This strengthens policy training, boosts real-world performance, and minimizes the sim-to-real gap, ultimately enhancing urban mobility and reducing congestion.

\section{Related Work}
\label{sec:relatedwork}
\subsection{Reinforcement Learning for Multi-Agent Traffic Signal Control}
Reinforcement Learning for multi-agent traffic signal control has emerged as a promising approach to alleviate urban traffic congestion by enabling intersections to operate as cooperative agents~\citep{choy2003cooperative}. Under this framework, each traffic signal controller is treated as an agent that learns optimal control policies through local interactions with the environment and limited communication with neighboring intersections~\citep{balaji2010multi}. Unlike traditional rule-based methods that rely on pre-defined heuristics~\citep{dion2002rule}, RL-based approaches dynamically adapt to real-time traffic conditions, yielding significant improvements in vehicle travel time and delay reduction~\citep{zheng2019diagnosing}. Multi-agent reinforcement learning (MARL) introduces both additional complexities and opportunities compared to single-agent settings~\citep{arel2010reinforcement}. Coordination among multiple agents can enhance overall network performance by balancing local decisions with global objectives, yet challenges such as environmental non-stationarity and the need for scalable communication strategies persist~\citep{chen2020toward}. Recent advances in MARL have explored solutions like centralized training with decentralized execution and cooperative learning schemes to overcome these challenges ~\citep{huang2021network}. Moreover, while many existing RL-based TSC methods focus on optimizing performance within simulated environments~\citep{mei2024libsignal}, the sim-to-real gap remains a critical hurdle~\citep{da2023sim2real}. Some recent studies have attempted to narrow this gap but only focus on the single-agent settings~\citep{da2023uncertainty, da2024prompt}, whereas our approach applies the work to more complex multi-agent settings, which hold great potential for more scalable TSC systems capable of effectively responding to dynamic traffic patterns.
\subsection{Sim-to-Real Methods for RL}
Sim-to-real transfer methods in RL broadly fall into three main categories~\citep{zhao2020sim, da2025survey}. The first category, \textbf{\emph{domain randomization}}~\citep{tobin2019real,andrychowicz2020learning,wei2022honor}, focuses on training policies that are robust to environmental variations by relying heavily on simulated data, which is particularly advantageous when facing uncertain or evolving target domains. The second category, \textbf{\emph{domain adaptation}}~\citep{tzeng2019deep,han2019learning}, addresses the challenge of distribution shifts between the source and target environments by aligning feature representations. Although many techniques in this category are aimed at bridging gaps in robotic perception~\citep{tzeng2015towards,fang2018multi,bousmalis2018using,james2019sim}, in the traffic signal control domain the discrepancy is mainly due to differences in dynamics, since most methods use vectorized observations such as lane-level vehicle counts or delays. The third category involves \textbf{\emph{grounding methods}}, which aim to reduce simulator bias and improve alignment with real-world dynamics. In contrast to system identification approaches~\citep{6907423,Cully_2015} that seek to learn exact physical parameters, Grounded Action Transformation (GAT)~\citep{hanna2017grounded} modifies the simulator dynamics via grounded actions, showing promising results for sim-to-real transfer in robotics~\citep{zhang2025reproducible}. Recent work~\citep{IROS20-Desai,IROS20-Karnan,NEURIPS20-Karnan} has further advanced grounding methods by incorporating stochastic modeling, reinforcement learning, and imitation-from-observation techniques. Our approach, \ours, builds on the GAT framework, introducing novel multi-agent designs and proposing local-joint solutions that consider neighbor state and action information.

\section{Preliminaries}
\label{sec:preliminaries}
This section introduces the necessary background for understanding our proposed method, including the formulation of the multi-agent reinforcement learning (MARL) traffic signal control (TSC) problem and an overview of Grounded Action Transformation (GAT). The detailed notation summary is shown in Table~\ref{tab:notations}.

\subsection{Multi-Agent Traffic Signal Control}
\label{subsec:multiAgentTsc}
We formulate TSC as a Decentralized Partially Observable Markov Decision Process (Dec-POMDP), where each intersection acts as an agent observing partial traffic states and optimizing local control policies to maximize cumulative reward. See Section~\ref{sup:decPOMDP} in the Supplementary Materials for full notation.

\subsection{Agent Design}
\label{sec:agentdesign}
For our agent design, we align with the most prevalent works in the TSC domain, such as PressLight~\citep{wei2019presslight}, with slight modifications, and use it consistently across all experiments. We summarize the state representation, action space, reward function, and learning method for our agents in Section~\ref{sec:agentDesignDetails} of the Supplementary Materials.

\subsection{Grounded Action Transformation}
\label{sec:vanillaGAT}
Grounded Action Transformation (GAT) is a framework designed to align simulated environments with real-world dynamics using real trajectories $\mathcal{D}_{\text{real}} = \{\tau^1, \dots, \tau^I\}$ collected by executing a policy $\pi_{\theta}$ in the real environment $E_{\text{real}}$. Let $P^*$ denote the real-world transition dynamics and $P_{\phi}$ denote the parameterized transition function of the simulator $E_{\text{sim}}$. GAT optimizes $\phi$ to minimize the discrepancy between $P^*$ and $P_{\phi}$:
\begin{equation}
\label{eq:klObjective}
    \phi^* = \arg\min_{\phi} \sum_{\tau^i \in \mathcal{D}_{\text{real}}} \sum_{t=0}^{T-1} d\left(P^*(s^i_{t+1} \mid s^i_t, a^i_t), P_{\phi}(s^i_{t+1} \mid s^i_t, a^i_t)\right),
\end{equation}
where $d(\cdot)$ is a distance measure (e.g., Kullback-Leibler divergence).

Given a policy $\pi_\theta$ that outputs an action $a_t$ to take in a given state $s_t$, GAT employs an action transformation function $g_{\phi}(s_t, a_t)$ parameterized by $\phi$ to compute a grounded action $\hat{a}_t$:

\begin{equation}
\label{eq:vanillaGAT}
    \hat{a}_t = g_{\phi}(s_t, a_t) = h_{\phi^{-}}\!\left(s_t, f_{\phi^{+}}(s_t, a_t)\right).
\end{equation}

The vanilla GAT framework consists of two models: a forward model $f_{\phi^{+}}$ and an inverse model $h_{\phi^{-}}$. The \textbf{forward model} takes as input the current state $s_t$ and the action $a_t$ from $E_{\text{sim}}$ and predicts the next state $\hat{s}_{t+1}$ in $E_{\text{real}}$: $\hat{s}_{t+1} = f_{\phi^{+}}(s_t, a_t)$. The \textbf{inverse model}, in turn, receives the current state $s_t$ from $E_{\text{sim}}$ and the predicted next state $\hat{s}_{t+1}$ from the forward model, generating a grounded action $\hat{a}_t$ that attempts to transition $s_t$ to $\hat{s}_{t+1}$ in $E_{\text{sim}}$: $\hat{a}_t = h_{\phi^{-}}(s_t, \hat{s}_{t+1})$. With effective grounding, the simulator’s transition dynamics, \( P_{\phi}\), better approximate those of the real environment, \(P^*\). This alignment facilitates more effective policy training in \( E_{\text{sim}} \), as GAT reduces the discrepancy in transition dynamics, leading to more realistic state transitions and ultimately reducing the sim-to-real gap. Note that the forward model $f_{\phi^{+}}$ is trained using data collected in $E_{\text{real}}$ and the inverse model $h_{\phi^{-}}$ is trained using data collected in $E_{\text{sim}}$.

\section{Grounded Action Transformation in Multi-Agent Settings}
\label{sec:gatInMulti}

\begin{figure}
    \centering
    \includegraphics[width=1\linewidth]{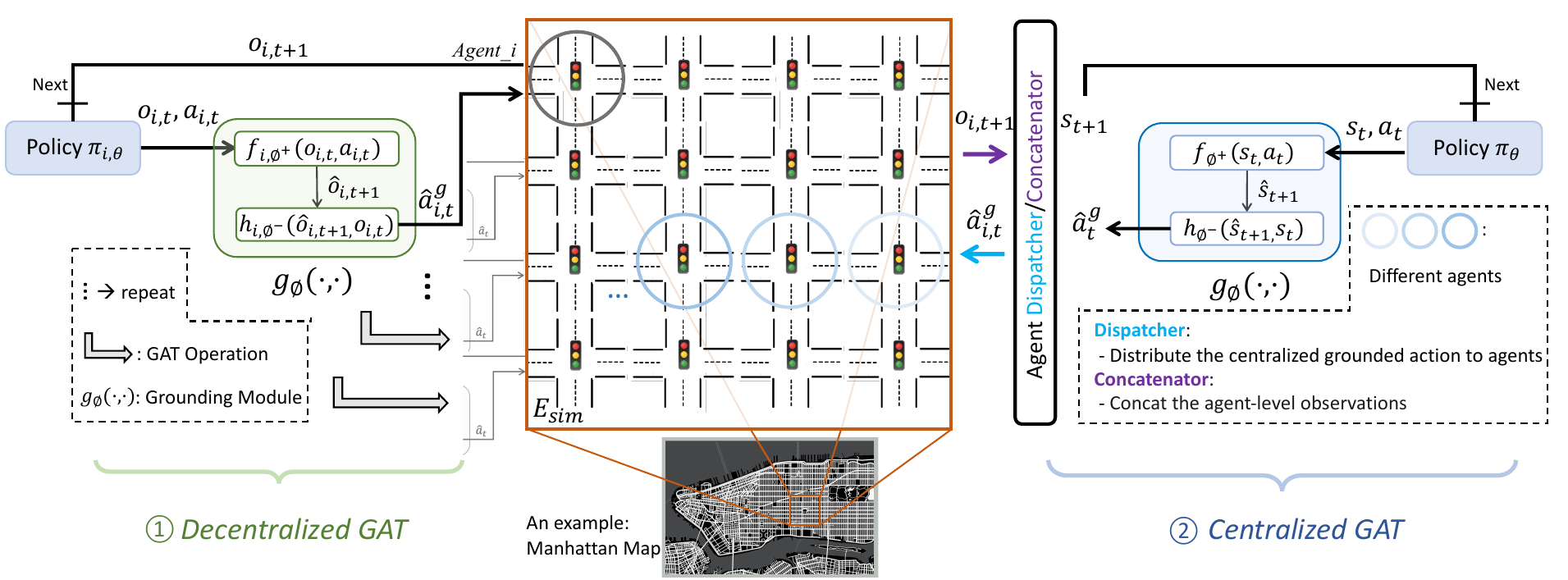}
    \caption{Overview of centralized and decentralized GAT in a 4×4 traffic network. \textbf{Left}: In decentralized GAT, each agent $i$ observes its state $o_{i,t}$ and selects an action $a_{i,t}$ via policy $\pi_{i,\theta}$. This action is passed through a forward model $f_{i,\phi^+}$ to predict the next observation $\hat{o}_{i,t+1}$, which is then input to the inverse model $h_{i,\phi^-}$ to produce the grounded action $\hat{a}^g_{i,t}$. This process runs independently for all agents. \textbf{Right}: Centralized GAT follows the same logic, but observations $o_{i,t}$ are concatenated into a global state $s_t$, and input to a shared forward model $f_{\phi^+}$ to produce $\hat{s}_{t+1}$ (a composition of all $\hat{o}_{i,t+1}$). This is then passed to a centralized inverse model $h_{\phi^-}$ to yield the global grounded actions $\hat{a}^g_t$, which are dispatched to agents to replace their original actions.
    }
    \label{fig:cenDecen}
\end{figure}

Grounded Action Transformation (GAT) bridges the sim-to-real gap using forward and inverse models to align simulator and real-world dynamics. In multi-agent settings, this alignment is challenged by inter-agent interactions. As shown in Figure~\ref{fig:cenDecen}, GAT can be centralized, capturing global dynamics at the cost of scalability, or decentralized, scaling well but ignoring multi-agent interactions. This section presents both approaches as a foundation for our proposed hybrid method, Joint-Local GAT (\ours) in Section~\ref{sec:jlgat}.

\subsection{Centralized Grounded Action Transformation}
A natural approach to multi-agent GAT is to treat the environment as a single-agent system by using global state and action inputs to a shared forward and inverse model. We provide an overview of centralized GAT in Figure~\ref{fig:cenDecen}. This enables the modeling of inter-agent dynamics but increases learning complexity as the number of agents grows. We retain the GAT objective from Equation~(\ref{eq:klObjective}), modifying it to use global states and actions. Following~\citep{da2024prompt}, we model $f_{\phi^+}$ and $h_{\phi^-}$ as neural networks trained via MSE and CCE losses. Unlike vanilla GAT, our inputs and outputs are global state-action tuples \(s_t, a_t\), composed of all agent observations \(o_{i,t}\) and actions \(a_{i,t}\).

\begin{itemize}
    \item The \textit{centralized forward model}, applied to traffic signal control, aims to predict the next global traffic state $\hat{s}_{t+1}$ in the real environment \(E_{\text{real}}\) after agents take global actions $a_{t}$ in the global traffic state $s_{t}$.
    
    \item The \textit{centralized inverse model}, applied to traffic signal control, considers the global traffic state $s_{t}$ in \(E_{\text{sim}}\) and predicted global next traffic state $\hat{s}_{t+1}$ in \(E_{\text{real}}\) from the forward model to predict global grounded actions $\hat{a}_{t}^{\text{g}}$. Note the inputs to the inverse model $h_{\phi^{-}}$ are global states and actions, but we compute CCE Loss to optimize $\phi^{-}$ by extracting the individual grounded actions $\hat{a}_{i,t}^{\text{g}}$ from the global grounded actions $\hat{a}_{t}^{\text{g}}$ and averaging across all agents for each sample.

\end{itemize}

\subsection{Decentralized Grounded Action Transformation}
\label{sec:decentralizedGAT}
A second intuitive approach to applying GAT to multi-agent settings is to assign each agent its own forward and inverse model. In this decentralized framework, each agent's GAT models operate independently, utilizing only their own information as if they were in a single-agent setting. We provide an overview of decentralized GAT in Figure~\ref{fig:cenDecen}. This improves scalability, allowing models to focus on local dynamics per agent. However, they ignore the influence of other agents, limiting their ability to model global dynamics. We follow~\citep{da2024prompt}, modifying inputs to use local observations in line with the Dec-POMDP formulation described in Section~\ref{subsec:multiAgentTsc}. Each agent $i$ learns its own $f_{i,\phi^+}$ and $h_{i,\phi^-}$ to model a local grounded transition function \(P_\phi\), still optimizing Equation~(\ref{eq:klObjective}) to minimize the discrepancy between $P^*$ and $P_{\phi}$.

\begin{itemize}
    \item The \textit{ decentralized forward model}, applied to traffic signal control, aims to predict the next state (observation) $\hat{o}_{t+1}$ of traffic in the real environment \(E_{\text{real}}\) for each agent $i$ after the action $a_{i,t}$ is taken in the current traffic observation $o_{i,t}$.
    
    \item The \textit{decentralized inverse model}, applied to traffic signal control, considers the traffic observation $o_{i,t}$ in $E_{\text{sim}}$ and the predicted next observation $\hat{o}_{i,t+1}$ in $E_{\text{real}}$ from the forward model to predict the grounded action $\hat{a}_{i,t}^{\text{g}}$ for each agent $i$.
    
\end{itemize}

\section{JL-GAT: Joint-Local Grounded Action Transformation}
\label{sec:jlgat}
\begin{figure}
    \centering
    \includegraphics[width=0.99\linewidth]{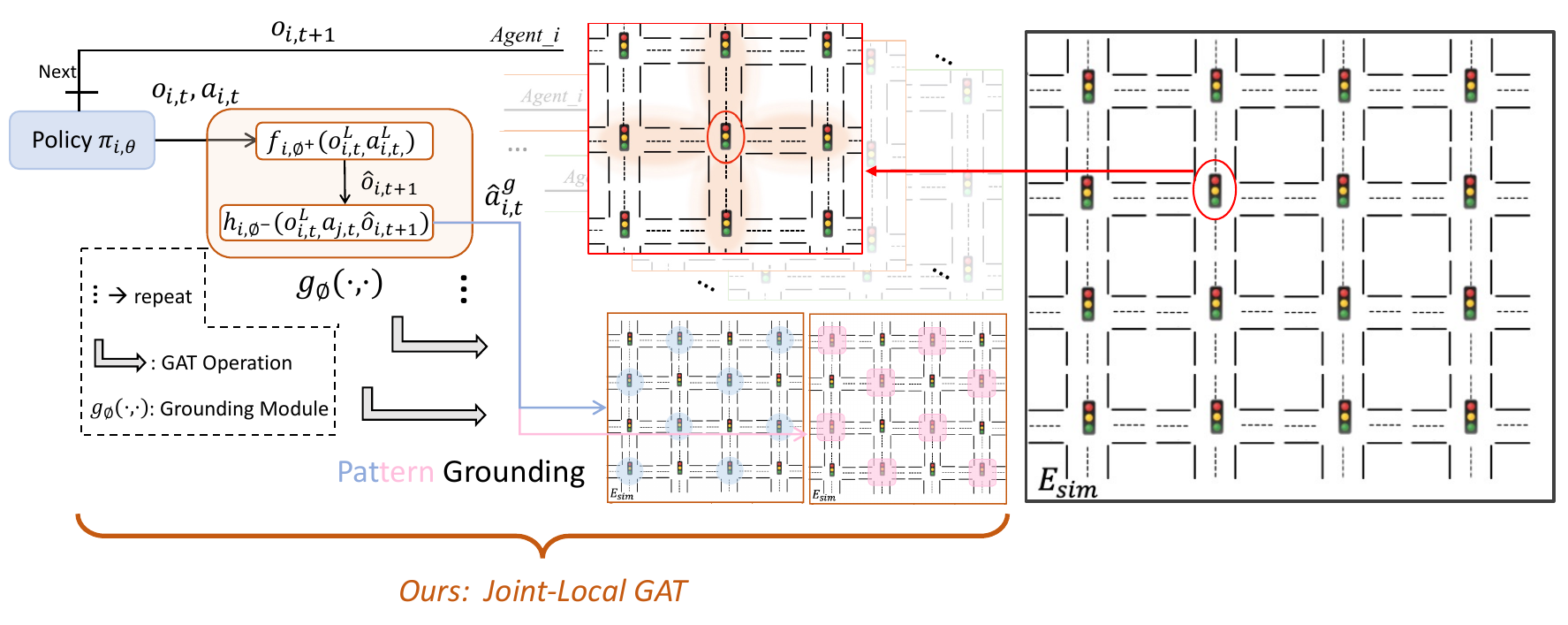}
    \caption{Overview of \ours. The pipeline follows these steps: Each agent $i$ first observes its state $o_{i,t}$ and selects an action $a_{i,t}$ using its policy $\pi_{i,\theta}$. The agent then incorporates neighboring agent observations and actions $o_{j,t}$, $a_{j,t}$ within a predefined sensing radius $r$, considering those within a Manhattan distance of $r$ or less. The 3×3 grid in the top center illustrates the neighboring information used for grounding when $r = 1$. The forward model $f_{i,\phi^{+}}$ takes in $o_{i,t}$, $a_{i,t}$ and neighboring $o_{j,t}$, $a_{j,t}$, forming the local joint observation $o^L_{i,t}$ and local joint action $a^L_{i,t}$. The forward model $f_{i,\phi^{+}}$ then predicts the next observation $\hat{o}_{i,t+1}$ for agent $i$. This predicted observation, along with the local joint observation $o^L_{i,t}$ and assumed neighboring actions $a_{j,t}$, is fed into the inverse model $h_{i,\phi^{-}}$. The inverse model $h_{i,\phi^{-}}$ outputs a grounded action $\hat{a}^g_{i,t}$ for agent $i$ to take instead of $a_{i,t}$. Finally, we address the cascading invalidation effect, a novel challenge arising with \ours, by introducing pattern grounding, illustrated in the bottom center.}
    \label{fig:mainmethod}
\end{figure}

By modifying our decentralized GAT formulation in Section \ref{sec:decentralizedGAT} to incorporate local joint state and action information for each agent, we arrive at \ours as shown in Figure~\ref{fig:mainmethod}. \ours strikes a balance between the two multi-agent applications of GAT, centralized and decentralized, introduced in Section \ref{sec:gatInMulti}. With this hybrid approach, \ours reaps unique benefits from both approaches, allowing GAT to be applied in large-scale multi-agent settings while still capturing essential agent interactions that influence the transition dynamics of the environment. 

\subsection{Overview of \ours}
We introduced two natural ways to apply GAT to multi-agent environments in Section \ref{sec:gatInMulti}: a centralized approach, which uses a single forward and inverse model to capture global information, and a decentralized approach, where each agent has its own GAT model, considering only its own state and actions. Although centralized GAT captures global interactions, it struggles to scale as the agent count grows. In contrast, decentralized GAT simplifies learning but ignores inter-agent dynamics that are critical to transition modeling. To overcome these limitations, we propose \ours, visualized in Figure~\ref{fig:mainmethod}. The core idea behind \ours is simple yet powerful: combine the strengths of both approaches by considering multi-agent interactions, such as in centralized GAT, while retaining the scalability of the decentralized approach. \ours achieves this by incorporating state and action information from neighboring agents into decentralized GAT models, preserving local agent interactions while maintaining the scalability of a decentralized setup. This results in more realistic simulated transitions, narrowing the sim-to-real gap.

\subsection{Formulation of \ours}
In this section, we formally define our proposed method, \ours. We first continue with the decentralized GAT approach described in Section \ref{sec:decentralizedGAT}, which includes a single forward and inverse model for each agent, extending it to reach the formulation of \ours. Then, we introduce the new objective for \ours. Lastly, we outline the forward and inverse model setup used in \ours, discussing the intuition behind the modifications and their benefits.

\subsubsection{\ours from Decentralized GAT}
\label{sec:extendDecGAT}
We build on the decentralized GAT formulation introduced in Section \ref{sec:decentralizedGAT}, where for each agent \(i\), we incorporate neighboring state and action information. We define the local joint state $o^{L}_{i,t}$ and action $a^{L}_{i,t}$ of agent \(i\) as its own observation $o_{i,t}$ and action $a_{i,t}$ at time $t$ combined with the observation and action information $o_{j,t}$, $a_{j,t}$ of agents $j$ within a predefined sensing radius $r$:

\[
o^{L}_{i,t} = \{ o_{i,t} \} \cup \{ o_{j,t} \mid d(i, j) \leq r \}, \quad a^{L}_{i,t} = \{ a_{i,t} \} \cup \{ a_{j,t} \mid d(i, j) \leq r \}
\]

where the Manhattan distance between agents \( i \) and \( j \) is defined as: $d(i, j) = |x_{i} - x_{j}| + |y_{i} - y_{j}|,$ with \( x_i, y_i \) and \( x_j, y_j\) representing the positions of agents \( i \) and \( j \) in a 2D coordinate space.

\subsubsection{Objective Function for \ours}
The formulation of \ours requires modifications to the objective in decentralized GAT shown in Equation (\ref{eq:decentralObjective}). Given real-world trajectories \( \mathcal{D}_{\text{real}} = \{\tau^1, \dots, \tau^I\} \), where each trajectory \( \tau^k = (s_t^k, a_t^k, s_{t+1}^k)_{t=0}^{T-1} \) is collected by executing policies in the real environment \(E_{\text{real}}\), our new objective is to learn a grounded simulator transition function \(P_{i,\phi}\) for each agent $i$ that minimizes:

\begin{equation}
\label{eq:decentralObjective}
    \phi^* = \arg\min_{\phi} \sum_{\tau^k \in \mathcal{D}_{\text{real}}} \sum_{t=0}^{T-1} d\left(P^*_i(o^{k}_{i,t+1} \mid o^{L,k}_{i,t}, a^{L,k}_{i,t}), P_{i,\phi}(o_{i,t+1}^{k} \mid o^{L,k}_{i,t}, a_{i,t}^{k,L})\right),
\end{equation}

where \(P^*_i\) represents real-world transition dynamics for an agent $i$ and \(d(\cdot)\) is a divergence measure (e.g., Kullback-Leibler divergence). We arrive at this objective by replacing the single-agent observations and actions from the vanilla GAT objective shown in Equation (\ref{eq:klObjective}) with local joint states (observations) and actions. Note that \ours attempts to model the transition to the next individual observation $o^k_{i,t+1}$ for a trajectory $k$ as opposed to a local joint observation.

\subsubsection{Forward and Inverse Models in \ours}
In this section, we present the forward and inverse models employed in \ours. We then highlight the advantages of our modifications to both vanilla and decentralized GAT. Finally, we explain how we strike a balance between centralized and decentralized GAT, effectively combining the strengths of both approaches.

$\bullet$~\textit{The forward model of \ours} predicts the next individual state $\hat{{o}}_{i,t+1}$ (observation) that would occur in the real environment \(E_{\text{real}}\) for agent $i$ if the local joint action $a^{L}_{i,t}$ was taken in local joint state $o^{L}_{i,t}$ at time $t$. Applied to traffic signal control, the forward model predicts the next real environment traffic state that would occur if the local joint action is taken in the current local joint traffic state:
    
\begin{equation}
\label{eq:jlgatForward}
    \hat{o}_{i,t+1} = f_{i,\phi^{+}}(o^{L}_{i,t}, a^{L}_{i,t})
\end{equation}
    
    Our setup of the forward model builds on that of the decentralized setup in Section \ref{sec:decentralizedGAT}, where we also approximate the forward model $f_{i,\phi^{+}}$ with a deep neural network for each agent $i$, now considering local joint information instead of only individual information, and optimize $\phi^{+}$ by minimizing the Mean Squared Error (MSE) loss:

\begin{equation}
\label{eq:jlgatForwardLoss}
    \mathcal{L}(\phi^+) = \text{MSE}({o}_{i,t+1}, \hat{{o}}_{i,t+1}) = \text{MSE}({o}_{i,t+1}, f_{i,\phi^{+}}(o^{L}_{i,t}, a^{L}_{i,t}))
\end{equation}

where $o^{L}_{i,t}$, $a^{L}_{i,t}$, and ${o}_{i,t+1}$ are sampled from trajectories collected in \(E_{\text{real}}\). Note that the forward model in \ours predicts a single next state (observation) $\hat{{o}}_{i,t+1}$ for each agent $i$ as in the decentralized GAT setup. In this way, \ours avoids the pitfall of attempting to predict neighboring agent observations, as those neighbors may be influenced by other agents at distance $d$ beyond the predefined radius $r$. Furthermore, by including the actions $a_{j,t}$ of neighboring agents $j$ within $r$, the forward model assumes that the neighboring agent actions will remain fixed. This assumption has significant implications for the setup of the inverse model in \ours, and if violated, gives way to the \textit{cascading invalidation effect} described in Section \ref{subsec:cascadingInvalidation}.

$\bullet$~\textit{The inverse model of \ours} predicts a grounded action $\hat{{a}}_{i,t}^{\text{g}}$ for agent $i$ at time $t$ that would attempt to transition the current local joint observation $o^{L}_{i,t}$ to the predicted individual next observation $\hat{o}_{i,t+1}$ in the simulated environment \(E_{\text{sim}}\). We further deviate from previous grounded action transformation works by including additional action information into the inverse model to predict a grounded action $\hat{{a}}_{i,t}^{\text{g}}$ for agent $i$. We use $a^{L}_{i,t}$, which incorporates the actions $a_{j,t}$ of neighboring agents $j$ within a predefined radius $r$ (as described in Section~\ref{sec:extendDecGAT}), as input to the inverse model for \ours. This implicitly assumes that their actions in $E_{\text{sim}}$ remain unchanged at time $t$:

    \begin{equation}
    \label{eq:jlgatInverse}
        \hat{{a}}_{i,t}^{\text{g}} = h_{i,\phi^{-}}(o^{L}_{i,t}, a^{L}_{i,t}, \hat{o}_{i,t+1})
    \end{equation}
    Including neighboring agent actions $a_{j,t}$ into the inverse model is invaluable for multi-agent settings, as it allows us to capture local agent interactions that affect the transition dynamics of a single agent $i$. Furthermore, we previously assumed neighboring agent actions would remain unchanged with our input to the forward model, thus it is a natural extension of the inverse model to also include this information. A key insight is that these assumptions lead to the \textit{cascading invalidation effect} described in Section~\ref{subsec:cascadingInvalidation}. We conduct an ablation study in Section~\ref{subsec:ablation}, on this additional information, further reinforcing its necessity in \ours. As in the forward model, we build on the inverse model from decentralized GAT in Section~\ref{sec:decentralizedGAT} and approximate $h_{i,\phi^{-}}$ with a deep neural network for each agent $i$ and optimize $\phi^{-}$ by minimizing the Categorical Cross-Entropy (CCE) Loss:

\begin{equation}
\label{eq:jlgatInverseLoss}
    \mathcal{L}(\phi^-) = CCE(a_{i,t}^{\text{g}}, \hat{a}_{i,t}^{\text{g}}) = CCE(a_{i,t}^{\text{g}}, h_{i,\phi^{-}}(o^{L}_{i,t}, a^{L}_{i,t}, \hat{o}_{i,t+1}))
\end{equation}

where $a_{i,t}^{\text{g}}$, $o^{L}_{i,t}$, and $\hat{o}_{i,t+1}$ are sampled from trajectories collected in \(E_{\text{sim}}\).

\subsection{Cascading Invalidation Effect}
\label{subsec:cascadingInvalidation}
While adapting \ours to include local joint information, we observe a unique challenge, namely the \textit{cascading invalidation effect}. This problem arises from the use of state and action information from neighboring agents to predict the next state that would occur in \(E_{\text{real}}\), as shown in Equation (\ref{eq:jlgatForward}). When using neighboring state and action information to attempt to bring the transition dynamics of \(E_{\text{sim}}\) closer to \(E_{\text{real}}\), the underlying assumption is that the actions of neighbor agents will remain unchanged in \(E_{\text{sim}}\). If the actions of an agent and one of its neighbors within the predefined radius $r$ are grounded simultaneously, both grounded actions become invalid and may no longer aid in reducing the sim-to-real gap. This is due to the fact that while grounding actions, we assume neighbor actions will not change. We also observe this effect cascade through a network of agents if grounding sequentially, as each agent grounds their action, assuming neighbor actions will remain unchanged. To overcome the cascading invalidation effect, we propose two different approaches:

$\bullet$~\textit{Pattern Grounding}. This approach is simple yet effective: we set a pattern to ground specific agents during a training epoch to avoid any grounding assumption conflicts. We visualize pattern grounding in Figure~\ref{fig:mainmethod}. For example, in our experiments for traffic signal control, we utilize a 1x3 traffic network and apply pattern grounding by grounding only the first and last agent for an epoch. Then, we ground only the agent in between them for the next epoch, alternating between the two set grounding patterns. This directly overcomes the cascading invalidation effect by avoiding grounding agents whose actions have been assumed fixed, but a rigid grounding pattern reduces flexibility during training. This approach can also be paired with \textit{probabilistic grounding}, but for our evaluations, we focused solely on applying each technique separately.
    
$\bullet$~\textit{Probabilistic Grounding}. In this approach, we let $P_{\text{ground}}^i(t)$ represent the probability of grounding an action $a_{i,t}$ for each agent $i$ at time step $t$:
    $P_{\text{ground}}^i(t) = p_{\text{ground}}$.    
Using probability to determine when grounding occurs introduces flexibility by allowing different grounding patterns to emerge naturally across epochs, as opposed to a fixed or rigid scheme. As demonstrated in Tables~\ref{tab:rainyTable} and~\ref{tab:snowyTable}, this approach led to strong performance for \ours. Although probabilistic grounding does not directly overcome the cascading invalidation effect as pattern grounding does, it often circumvents this challenge by using a fixed probability to ground, which introduces some trade-offs. In particular, this can lead to training scenarios in the simulated environment $E_{\text{sim}}$ that do not accurately reflect the transition dynamics of the real environment $E_{\text{real}}$. This is due to the less restrictive grounding requirements in probabilistic grounding compared to pattern grounding, which enables agents to ground their actions independently without requiring consideration of whether neighboring agents are simultaneously utilizing their actions for grounding. Furthermore, decreasing the grounding probability $P_{\text{ground}}^i(t)$ for each agent $i$ inherently mitigates the likelihood of cascading invalidation. However, this comes at the cost of reducing the amount of grounding during training, which may result in a larger sim-to-real gap. We experiment with various probabilities in Section~\ref{sec:probability}, where we recommend $1/N$ as a starting point for probabilistic grounding based on empirical evaluation.

We acknowledge that there are several alternative solutions to the cascading invalidation effect that remain to be explored, such as clustering groups for grounding, learned grounding patterns, and algorithmic approaches to grounding. These avenues are left for future work.

\subsection{Training Algorithm}
We present the training procedure for \ours in Algorithm~\ref{alg:jlgatAlgo}. The algorithm takes as input initial policies $\pi_{i,\theta}$, forward models $f_{i,\phi^{+}}$, and inverse models $h_{i,\phi^{-}}$ for each agent $i$, as well as simulation and real-world datasets $\mathcal{D}_{\text{sim}}$ and $\mathcal{D}_{\text{real}}$ (collected offline or from rollouts~\citep{da2023uncertainty}). A sensing radius $r$ is required to determine neighboring agent interactions for grounding, and an optional grounding pattern or probability may also be specified. The output includes updated policies and models for each agent. Training begins with $M$ iterations of policy pre-training in \(E_{\text{sim}}\), followed by multiple epochs consisting of: (1) optional policy rollouts in \(E_{\text{sim}}\) and \(E_{\text{real}}\) to populate $\mathcal{D}_{\text{sim}}$ and $\mathcal{D}_{\text{real}}$; (2) updates to $f_{i,\phi^{+}}$ and $h_{i,\phi^{-}}$ using the collected data; (3) policy training episodes using grounded actions to align simulated dynamics with the real world; and (4) reinforcement learning-based policy updates in \(E_{\text{sim}}\) with improved dynamics to reduce the sim-to-real gap.

\section{Experiments and Results}
In this section, we introduce our experiment setup and evaluation metrics, which closely follow that of~\citep{da2024prompt}, demonstrating both the existence of a performance gap between simulation and real environments and the effectiveness of \ours in reducing this gap. We also perform an ablation study to demonstrate the necessity of all additional information to the forward and inverse models in \ours. Lastly, we evaluate different probabilistic grounding settings and explore pairing \ours with the uncertainty quantification method from~\citep{da2023uncertainty}.

\subsection{Environments}
\label{sec:environments}
We built our implementation of \ours on top of LibSignal~\citep{mei2024libsignal}, an open-source environment for traffic signal control with multiple simulation environments. For our experiments, we consider CityFlow~\citep{zhang2019cityflow} as the simulation environment \(E_{\text{sim}}\), and SUMO~\citep{behrisch2011sumo} as the real environment \(E_{\text{real}}\). We use a sim-to-sim setup to mimic a sim-to-real deployment process with the main benefit of reproducibility and avoiding the negative impact of unexpected behaviors in the real world, as described in~\citep{da2024prompt, da2024probabilistic}. Our experiments consider two environmental conditions to showcase the sim-to-real gap: rainy and snowy, and we detail their parameter settings in Table~\ref{tab:environmentSettings} (Supplementary Materials).

\begin{itemize}
    \item \textit{Default settings}. This represents the default settings for CityFlow and SUMO, which we consider \(E_{\text{sim}}\) and \(E_{\text{real}}\), respectively.
    \item \textit{Adverse Weather conditions}. We model the effect of adverse weather conditions that are unaccounted for when training a TSC policy in \(E_{\text{sim}}\) by varying parameters in \(E_{\text{real}}\), such as acceleration, deceleration, emergency deceleration, and startup delay shown in Table \ref{tab:environmentSettings}. We attempt to mimic real-world adverse weather effects, such as wet and icy roads, by reducing the acceleration and deceleration rates of vehicles and increasing their startup delay.
\end{itemize}

\subsection{Evaluation Metrics}
Building on common practices in traffic signal control (TSC), as described in recent literature~\citep{wei2021recent}, we adopt the following standard metrics to assess policy performance. Average Travel Time (ATT) represents the average travel time $t$ for vehicles in a given road network, where lower ATT values indicate better control policy performance. Queue measures the number of vehicles waiting at a particular intersection, and we report the average queue over all intersections in a given road network, with smaller values being preferable. Delay captures the average time $t$ that vehicles wait in the traffic network, where lower delay is desirable. Throughput (TP) quantifies the number of vehicles that have completed their trip in a given road network, with higher TP values being better. Lastly, reward represents the return associated with taking an action $a_t$ in a state $s_t$ in RL. We use the same reward metric as~\citep{wei2019presslight}, defining the reward as negative pressure, and we report the sum of rewards for all intersections in our experiments.

In this work, we adopt the calculation metric for the performance gap between \(E_{\text{sim}}\) and \(E_{\text{real}}\) from~\citep{da2024prompt} and ~\citep{da2023uncertainty}. Specifically, for a metric $\psi$, we use the following equation to calculate the gap $\Delta$: $\psi_{\Delta} = \psi_{\text{real}} - \psi_{\text{sim}}$. Our goal is to reduce this sim-to-real gap by bringing the transition dynamics of \(E_{\text{sim}}\) closer to \(E_{\text{real}}\) while training through GAT. We report the $\Delta$ values for each metric, where smaller values are better for $\textit{ATT}_\Delta$, $\textit{Queue}_\Delta$, and $\textit{Delay}_\Delta$, and larger values are better for $\textit{TP}_\Delta$, and $\textit{Reward}_\Delta$ because they are negative values.

\subsection{Main Results}

To highlight the sim-to-real gap in multi-agent traffic signal control (TSC), we conduct experiments in the rainy and snowy settings introduced in Section~\ref{sec:environments}, with parameters detailed in Table~\ref{tab:environmentSettings}. We begin by evaluating the Direct Transfer approach: agents (Section~\ref{sec:agentdesign}) are trained from scratch in \(E_{\text{sim}}\) through six independent trials of 300 epochs each. After six independent trials for each network size, the best-performing policies based on lowest average travel time (ATT) are tested in \(E_{\text{real}}\). These policies then serve as initialization for various GAT-based multi-agent training configurations, including \ours. The resulting performance metrics are visualized in Figures~\ref{fig:mainResults} and~\ref{fig:mainResults2} in the Supplementary Materials. Full numerical results, including standard deviations and sim-to-real gap calculations, are provided in Tables~\ref{tab:rainyTable} and~\ref{tab:snowyTable}. A clear drop in performance is observed when directly transferring policies from \(E_{\text{sim}}\) to \(E_{\text{real}}\), illustrating the gap between simulation and real environments.

\begin{table}[htbp]
    \caption{Rainy environment performance using Direct Transfer as compared to centralized GAT, decentralized GAT, and two versions of our proposed method \ours. For each GAT configuration and network size pair, we run six independent trials and identify the best epoch in each trial based on the lowest average travel time (ATT) in \(E_{\text{real}}\). Reported metrics are averaged across these six best epochs (one per trial). The value in () shows the metric gap $\psi$ between \(E_{\text{sim}}\) and \(E_{\text{real}}\) and $\pm$ shows the sample standard deviation after six trials. The $\uparrow$ indicates that a higher value represents a better performance for a metric and the $\downarrow$ indicates that a lower value represents a better performance for a metric. Note that Direct Transfer is reported as the policies from the best performing epoch (by lowest ATT) in \(E_{\text{sim}}\) being tested in \(E_{\text{real}}\) after six trials of 300 epochs.} 
    \centering
    \resizebox{\textwidth}{!}{ 
        \begin{tabular}{c|cccccc}
            \toprule
            \bf Network & \bf Method & \bf ATT ($\Delta\downarrow$) & \bf Queue ($\Delta\downarrow$) & \bf Delay ($\Delta\downarrow$) & \bf TP ($\Delta\uparrow$) & \bf Reward ($\Delta\uparrow$)\\
            \hline
            \multirow{5}{*}{\bf 1x3} & Direct Transfer & 309.90 (188.64) & 67.66 (43.60) & 0.64 (0.23) & 4784 (-776) & -202.85 (-141.21) \\
            & Centralized GAT & 297.57(176.31)$\pm$16.12 & 65.59(41.53)$\pm$5.83 & 0.63(0.22)$\pm$0.01 & 4857(-702)$\pm$96.37 & -189.87(-128.23)$\pm$15.64 \\
            & Decentralized GAT & 276.92(155.66)$\pm$16.53 & 59.07(35.01)$\pm$6.00 & 0.63(0.22)$\pm$0.01 & 5004(-555)$\pm$119.40 & -175.30(-113.66)$\pm$16.12 \\
            & JL-GAT (Pattern) & 265.64(144.38)$\pm$6.72 & 51.96(27.90)$\pm$3.52 & 0.62(0.21)$\pm$0.005 & \textbf{5073(-487)$\pm$39.36} & -156.27(-94.63)$\pm$10.10 \\
            & JL-GAT (Probabilistic $1/N$ = 33\%) & \textbf{263.01(141.75)$\pm$2.59} & \textbf{50.63(26.57)$\pm$1.76} & \textbf{0.61(0.20)$\pm$0.005} & 5065(-494)$\pm$39.77 & \textbf{-152.12(-90.48)$\pm$5.08} \\
            \hline
            \multirow{5}{*}{\bf 4x4} & Direct Transfer & 485.63(158.38) & 6.89(5.39) & 0.19(0.11) & 2608(-320) & -90.77(-71.48) \\
            & Centralized GAT & 485.63(158.38)$\pm$0.00 & 6.89(5.39)$\pm$0.00 & 0.19(0.11)$\pm$0.00 & 2608(-320)$\pm$0.00 & -90.77(-71.48)$\pm$0.00 \\
            & Decentralized GAT & 476.69(149.44)$\pm$4.53 & 6.39(4.88)$\pm$0.37 & 0.18(0.10)$\pm$0.004 & 2620(-307)$\pm$10.30 & -84.31(-65.03)$\pm$2.38 \\
            & JL-GAT (Pattern) & 468.81(141.56)$\pm$2.42 & 5.99(4.48)$\pm$0.12 & \textbf{0.18(0.10)$\pm$0.002} & \textbf{2627(-300)$\pm$5.05} & \textbf{-83.47(-64.18)$\pm$2.08} \\
            & JL-GAT (Probabilistic $1/N$ = 6.25\%) & \textbf{467.11(139.86)$\pm$1.77} & \textbf{5.85(4.34)$\pm$0.17} & 0.18(0.10)$\pm$0.004 & 2625(-302)$\pm$7.06 & -85.33(-66.04)$\pm$1.60 \\
            \bottomrule
        \end{tabular}
         }
    \label{tab:rainyTable}
\end{table}

\begin{table}[htbp]
    \caption{Snowy environment performance using Direct Transfer as compared to centralized GAT,
 decentralized GAT, and two versions of our proposed method \ours. Refer to Table~\ref{tab:rainyTable} for details on the reporting methodology.}
    \centering
    \resizebox{\textwidth}{!}{ 
        \begin{tabular}{c|cccccc}
            \toprule
            \bf Network & \bf Method & \bf ATT ($\Delta\downarrow$) & \bf Queue ($\Delta\downarrow$) & \bf Delay ($\Delta\downarrow$) & \bf TP ($\Delta\uparrow$) & \bf Reward ($\Delta\uparrow$)\\
            \hline
            \multirow{5}{*}{1x3} & Direct Transfer & 473.29 (352.02) & 49.11 (25.05) & 0.66 (0.24) & 4297 (-1263) & -160.69 (-99.05) \\
            & Centralized GAT & 472.67(351.41)$\pm$1.51 & 49.20(25.14)$\pm$0.21 & \textbf{0.65(0.24)$\pm$0.004} & 4316(-1243)$\pm$47.77 & -160.46(-98.82)$\pm$0.57 \\
            & Decentralized GAT & 463.37(342.11)$\pm$11.84 & 54.27(30.21)$\pm$7.52 & 0.66(0.25)$\pm$0.01 & 4402(-1157)$\pm$96.01 & -166.85(-105.21)$\pm$17.78 \\
            & JL-GAT (Pattern) & 459.28(338.02)$\pm$2.79 & 50.59(26.53)$\pm$5.12 & 0.65(0.24)$\pm$0.01 & 4414(-1145)$\pm$40.40 &-157.20(-95.56)$\pm$11.17 \\
            & JL-GAT (Probabilistic $1/N$ = 33\%) & \textbf{456.14(334.88)$\pm$6.09} & \textbf{46.39(22.33)$\pm$3.26} & 0.65(0.24)$\pm$0.005 & \textbf{4436(-1123)$\pm$27.18} & \textbf{-147.97(-86.33)$\pm$9.20} \\
            \hline
            \multirow{5}{*}{4x4} & Direct Transfer & 593.06 (265.81) & 6.83 (5.33) & 0.20 (0.12) & 2423 (-505) & -96.28 (-76.99) \\
            & Centralized GAT & 593.06(265.81)$\pm$0.00 & 6.83(5.33)$\pm$0.00 & 0.20(0.12)$\pm$0.00 & 2423(-505)$\pm$0.00 & -96.28(-76.99)$\pm$0.00 \\
            & Decentralized GAT & 573.07(245.82)$\pm$4.09 & 5.70(4.19)$\pm$0.27 & 0.19(0.11)$\pm$0.004 & 2467(-460)$\pm$4.97 & -83.90(-64.61)$\pm$3.17 \\
            & JL-GAT (Pattern) & 567.75(240.50)$\pm$1.96 & 5.50(3.99)$\pm$0.08 & 0.19(0.11)$\pm$0.005 & \textbf{2471(-457)$\pm$7.85} & -83.83(-64.54)$\pm$1.51 \\
            & JL-GAT (Probabilistic $1/N$ = 6.25\%) & \textbf{566.22(238.97)$\pm$2.64} & \textbf{5.28(3.77)$\pm$0.18} & \textbf{0.18(0.10)$\pm$0.004} & 2470(-457)$\pm$3.97 & \textbf{-82.32(-63.03)$\pm$1.24} \\
            \bottomrule
        \end{tabular}
    }
    \label{tab:snowyTable}
\end{table}

\subsection{Ablation Study}
\label{subsec:ablation}
\begin{figure}[htbp]
    \centering
    \includegraphics[width=0.99\linewidth]{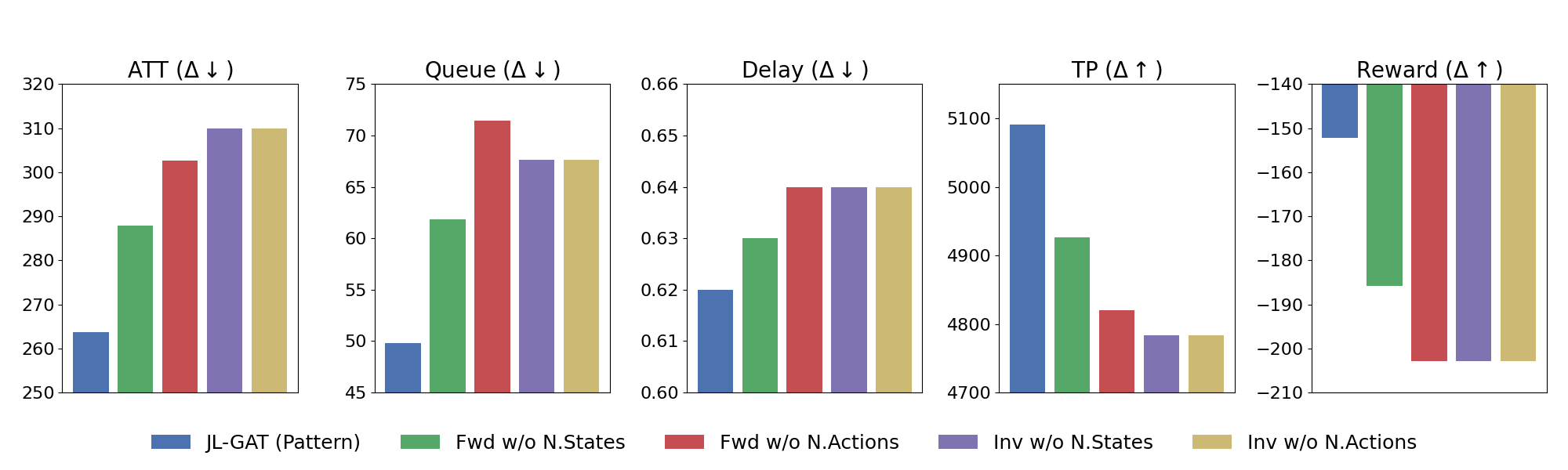}
    \caption{The ablation study on the proposed method. We systematically vary the information used in the GAT models of \ours to demonstrate the necessity of including neighboring agent information in all parts of GAT. The bars show the average performance of each metric over the best episodes of each method after two trials in the 1x3 rainy environment. Each plot displays the methods in the order they appear from left to right, as indicated in the legend. Full details including sim-to-real gap computation and sample standard deviation are shown in Table~\ref{tab:ablationTable}.}
    \label{fig:ablation}
\end{figure}

To show how different parts in \ours help sim-to-real transfer, we conduct an ablation study on the addition of neighboring information in the forward and inverse models of \ours. For this study, we focus on the rainy 1x3 environment while systematically varying the removal of neighboring states and action information used in \ours. We present the average performance of each metric for the best episode of each method. These results are based on two trials over 300 epochs, as shown in Figure~\ref{fig:ablation}, with full details including sim-to-real gap computation and sample standard deviation shown in Table~\ref{tab:ablationTable}. The last two methods failed to improve the Direct Transfer models used for initialization, indicating the necessity of all required modules for \ours.

\subsection{Probabilistic Grounding Settings}
\label{sec:probability}
We experiment with various probability grounding settings for \ours to test the robustness of \ours for different probability settings. We focus on four different variations of probability grounding, including $1/N$, which sets the grounding probability proportional to the number of agents in the environment. We report the best performance for each setting over 300 epochs in Table \ref{tab:probTable}. The results show that using a probability of 0.2 produces the best performance across all metrics in the 1x3 rainy environment. However, we recommend $1/N$ as a starting place for probabilistic grounding, as our results from Tables~\ref{tab:rainyTable},~\ref{tab:snowyTable}, and~\ref{tab:probTable} demonstrate a strong performance from the $1/N$ setting.

\begin{table}[htbp]
    \caption{Probability grounding settings for \ours in 1x3 rainy environment.}
    \centering
    \resizebox{0.9\textwidth}{!}{ 
        \begin{tabular}{cccccc}
            \toprule
            \bf Probability & \bf ATT ($\Delta\downarrow$) & \bf Queue ($\Delta\downarrow$) & \bf Delay ($\Delta\downarrow$) & \bf TP ($\Delta\uparrow$) & \bf Reward ($\Delta\uparrow$)\\
            \hline
            0.2 & \textbf{260.77(139.51)$\pm$4.73} & \textbf{50.23(26.17)$\pm$2.24} & 0.62(0.21)$\pm$0.005 & \textbf{5115(-445)$\pm$36.06} & \textbf{-151.34(-89.69)$\pm$5.09} \\
            0.5 & 281.73(160.47)$\pm$29.87 & 56.19(32.14)$\pm$16.36 & \textbf{0.61(0.20)$\pm$0.01} & 4909(-651)$\pm$209.30 & -170.52(-108.87)$\pm$39.52 \\
            0.8 & 297.75(176.49)$\pm$6.70 & 66.78(42.73)$\pm$5.97 & 0.63(0.22)$\pm$0.0001 & 4828(-732)$\pm$276.48 & -187.69(-126.05)$\pm$7.32 \\
            $1/N$ (0.3) & 261.56(140.30)$\pm$1.30 & 50.28(26.22)$\pm$2.59 & \textbf{0.61(0.20)$\pm$0.01} & 5062(-498)$\pm$25.38 & -155.33(-93.68)$\pm$4.24 \\
            \bottomrule
        \end{tabular}
        
    }
    \label{tab:probTable}
\end{table}

\subsection{\ours with Uncertainty Quantification}
Sim-to-real transfer can introduce uncertainty in action effectiveness due to discrepancies between simulated and real-world dynamics. Prior work has investigated uncertainty quantification (UQ) techniques to improve the reliability of decision-making~\citep{abdar2021review, liu2025uncertainty} such as MC dropout~\citep{gal2016dropout}, Deep Ensembles~\citep{rahaman2021uncertainty}, Evidential Deep Learning (EDL)~\citep{deng2023uncertainty}, and methods based on eigenvalues~\citep{thompson2019eigenvector}, etc. To explore whether UQ can enhance \ours, we incorporate the dynamic grounding rate method from~\citep{da2023uncertainty}, which adjusts the application of grounding based on model uncertainty. Specifically, for each agent, we compute the average model uncertainty over the previous two epochs and use it to determine whether to ground the current action. If the agent’s predicted uncertainty exceeds a dynamic threshold, the original (non-grounded) action is used instead. We evaluate this uncertainty-aware version of \ours in both rainy and snowy environments over three trials of 300 epochs each, and display the results in Table~\ref{tab:uncertaintyTable}. The results indicate that integrating UQ with \ours further reduces the sim-to-real gap in the 1×3 setting.

\begin{table}[htbp]
    \caption{Uncertainty quantification in \ours for 1x3 traffic network.}
    \centering
    \resizebox{\textwidth}{!}{ 
        \begin{tabular}{c|cccccc}
            \toprule
            \bf Environment & \bf Method & \bf ATT ($\Delta\downarrow$) & \bf Queue ($\Delta\downarrow$) & \bf Delay ($\Delta\downarrow$) & \bf TP ($\Delta\uparrow$) & \bf Reward ($\Delta\uparrow$)\\
            \hline
            \multirow{2}{*}{\bf Rainy} & JL-GAT (Pattern) & 263.61(142.35)$\pm$4.66 & 49.82(25.76)$\pm$1.46 & \textbf{0.62(0.21)$\pm$0.004} & 5091(-469)$\pm$20.26 & -152.20(-90.55)$\pm$5.96 \\
            & \ours w/ Uncertainty & \textbf{261.53(140.26)$\pm$4.56} & \textbf{49.65(25.59)$\pm$4.19} & \textbf{0.62(0.21)$\pm$0.01} & \textbf{5092(-468)$\pm$16.07} & \textbf{-148.15(-86.51)$\pm$11.73} \\
            \hline
            \multirow{2}{*}{\bf Snowy}& JL-GAT (Pattern) & 459.46(338.20)$\pm$3.89 & 47.13(23.07)$\pm$4.56 & 0.65(0.24)$\pm$0.01 & 4417(-1143)$\pm$20.26 & -150.40(-88.76)$\pm$12.10 \\
            & \ours w/ Uncertainty & \textbf{456.92(335.66)$\pm$4.87} & \textbf{44.51(20.45)$\pm$8.23} & \textbf{0.64(0.23)$\pm$0.02} & \textbf{4444(-1116)$\pm$48.87} & \textbf{-141.41(-79.76)$\pm$15.80} \\
            \bottomrule
            
        \end{tabular}
    }
    \label{tab:uncertaintyTable}
\end{table}

\section{Conclusion}
We have identified a significant performance gap that arises when directly transferring MARL-based TSC policies from simulation to the real world, primarily due to shifts in environment dynamics. To address this, we proposed \ours, a scalable framework that extends Grounded Action Transformation (GAT) to the MARL-based TSC setting. \ours enhances the performance of a decentralized approach to GAT, where each agent has its own GAT models, by incorporating neighboring agent information. This allows \ours to model inter-agent dynamics as in a centralized approach, without sacrificing the scalability of a decentralized approach. Extensive experiments across diverse traffic networks and simulated adverse weather conditions confirm that the hybrid design of \ours consistently reduces the sim-to-real performance gap. A key challenge we identified in the multi-agent GAT setting is the \textit{cascading invalidation effect}, which arises when multiple agents simultaneously ground their actions under the incorrect assumption that neighboring agents' actions remain fixed. Although we introduced two methods to mitigate this issue, a promising direction for future work lies in dynamically selecting which agents should engage in GAT and when.

\section*{Acknowledgments}
The work was partially supported by NSF awards \#2421839, Amazon Research Awards, NAIRR \#240120 and used AWS through the CloudBank project, which is supported by NSF grant \#1925001. The views and conclusions in this paper are those of the authors and should not be interpreted as representing any funding agencies.

\bibliographystyle{unsrtnat}
\bibliography{references}  






\appendix

\section{Additional Tables and Figures}

\begin{table}[H]
    \caption{Environment settings used in all experiments.}
    \centering
    \resizebox{0.7\textwidth}{!}{ 
        \begin{tabular}{ccccc}
            \hline
            \bf Environment & \bf Accel ($m/s^2$) & \bf Decel ($m/s^2$) & \bf E. Decel ($m/s^2$) & \bf S. Delay ($s$)\\
            \hline
            Default (\(E_{\text{sim}}\)) & 2.0 & 4.5 & 9.0 & 0.0 \\
            Rainy & 0.75 & 3.5 & 4.0 & 0.25 \\
            Snowy & 0.5 & 1.5 & 2.0 & 0.5 \\
            \hline
        \end{tabular}
    }
    \label{tab:environmentSettings}
\end{table}

\begin{table}[H]
    \caption{Ablation Study of \ours in 1x3 Rainy Environment.}
    \centering
    \resizebox{\textwidth}{!}{ 
        \begin{tabular}{cccccc}
            \toprule
            \bf Method & \bf ATT ($\Delta\downarrow$) & \bf Queue ($\Delta\downarrow$) & \bf Delay ($\Delta\downarrow$) & \bf TP ($\Delta\uparrow$) & \bf Reward ($\Delta\uparrow$)\\
            \hline
            JL-GAT (Pattern) & \textbf{263.61(142.35)$\pm$4.66} & \textbf{49.82(25.76)$\pm$1.46} & \textbf{0.62(0.21)$\pm$0.004} & \textbf{5091(-469)$\pm$20.26} & \textbf{-152.20(-90.55)$\pm$5.96} \\
            Forward Model w/o Neigh. States & 287.96(166.70)$\pm$31.03 & 61.82(37.76)$\pm$8.26 & 0.63(0.22)$\pm$0.01 & 4926(-634)$\pm$201.53 & -185.76(-124.11)$\pm$24.18 \\
            Forward Model w/o Neigh. Actions & 302.65(181.38)$\pm$10.26 & 71.41(47.36)$\pm$5.30 & 0.64(0.23)$\pm$0.01 & 4820(-740)$\pm$50.91 & -202.86(-141.22)$\pm$0.01 \\
            Inverse Model w/o Neigh. States & 309.90(188.64)$\pm$0.00 & 67.66(43.60)$\pm$0.00 & 0.64(0.23)$\pm$0.00 & 4784(-776)$\pm$0.00 & -202.85(-141.21)$\pm$0.00 \\
            Inverse Model w/o Neigh. Actions & 309.90(188.64)$\pm$0.00 & 67.66(43.60)$\pm$0.00 & 0.64(0.23)$\pm$0.00 & 4784(-776)$\pm$0.00 & -202.85(-141.21)$\pm$0.00 \\
            \bottomrule
        \end{tabular}
        }
    \label{tab:ablationTable}
\end{table}

\begin{table}[H]
    \caption{Key Notations and Descriptions in This Paper.}
    \begin{center}
        \begin{tabular}{ll}
        \toprule
            \multicolumn{1}{l}{\bf Symbol}  &\multicolumn{1}{l}{\bf Description}
            \\ \midrule
            $\mathcal{N}$ & Set of agents (traffic signals) \\
            $\mathcal{S}$ & Global state space \\
            $\mathcal{A}_i$ & Action space for agent $i$ \\
            $P$ & Transition function \\
            $R$ & Reward function \\
            $\gamma$ & Discount factor \\
            $o_{i,t}$ & State (observation) of agent $i$ at time $t$ \\
            $a_{i,t}$ & Action of agent $i$ at time $t$ \\
            $\hat{o}_{i,t+1}$ & Predicted next state (observation) for agent $i$ \\
            $\pi_i$ & Policy of agent $i$ \\
            $J_i$ & Expected cumulative reward for agent $i$ \\
            $\mathcal{D}_{\text{real}}$ & Real-world trajectory dataset \\
            $\mathcal{D}_{\text{sim}}$ & Simulation trajectory dataset \\
            $P^*$ & Real-world transition dynamics \\
            $P_{\phi}$ & Parameterized simulator dynamics \\
            $f_{i,\phi^{+}}$ & Forward model for agent $i$\\
            $h_{i,\phi^{-}}$ & Inverse model for agent $i$\\
            $r$ & Sensing radius \\
            $d(i, j)$ & Distance between agents \(i\) and \(j\) \\
            $s_{t}, a_{t}$ & Global state and action at time $t$ \\
            $o^{L}_{i,t}, a^{L}_{i,t}$ & Local joint state (observations) and actions for agent $i$ at time $t$ \\
            $\hat{a}_{t}^{\text{g}}$ & Global grounded action at time $t$ \\
            $\hat{{a}}_{i,t}^{\text{g}}$ & Grounded action for agent $i$ at time $t$ \\ \bottomrule
            
        \end{tabular}
    \end{center}
    \label{tab:notations}
\end{table}

\begin{figure}[H]
    \centering
    \includegraphics[width=0.99\linewidth]{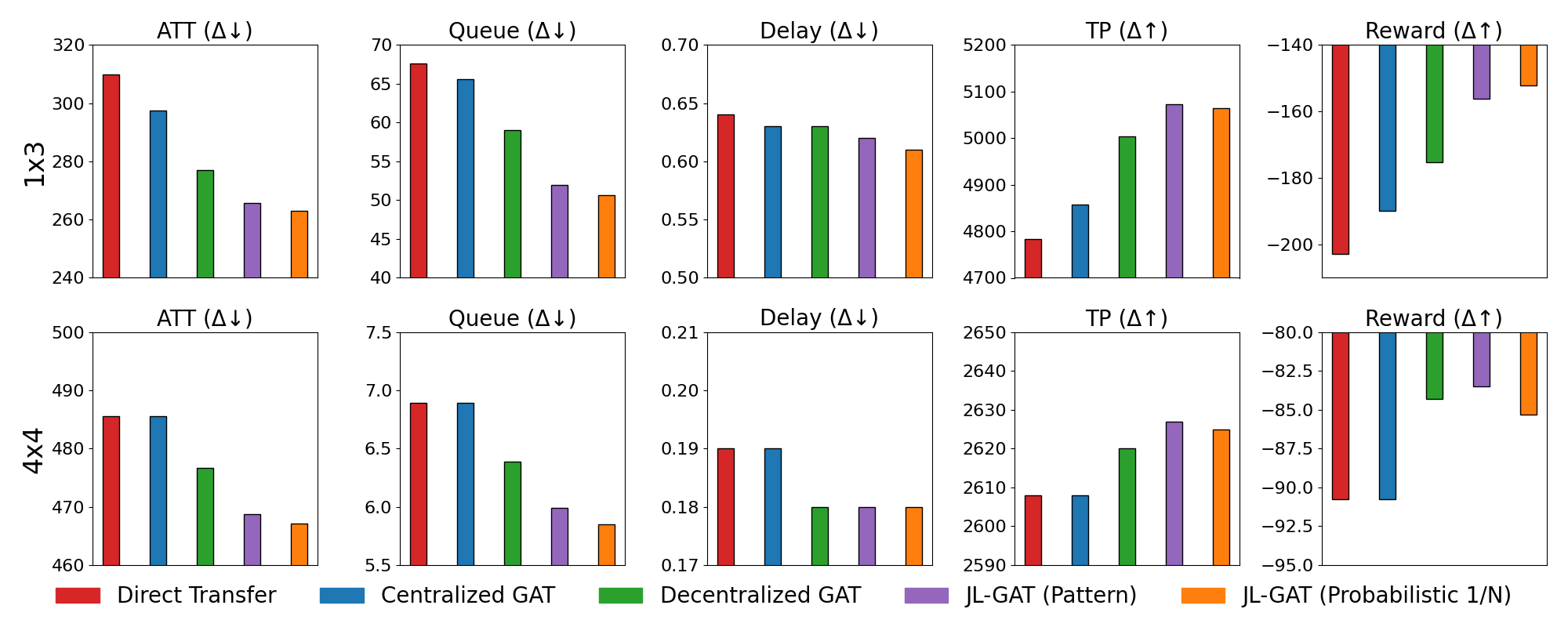}
    \caption{Average performance metrics over the best episode from each trial in the \textbf{rainy} environment. Top row: 1×3 traffic network. Bottom row: 4×4 traffic network. The $\uparrow$ indicates that a higher value represents a better performance for a metric and the $\downarrow$ indicates that a lower value represents a better performance for a metric. Each plot displays the methods in the order they appear from left to right, as indicated in the legend. Full quantitative results, including standard deviations and sim-to-real gap values, are presented in Table~\ref{tab:rainyTable}.}
    \label{fig:mainResults}
\end{figure}

\begin{figure}[H]
    \centering
    \includegraphics[width=0.99\linewidth]{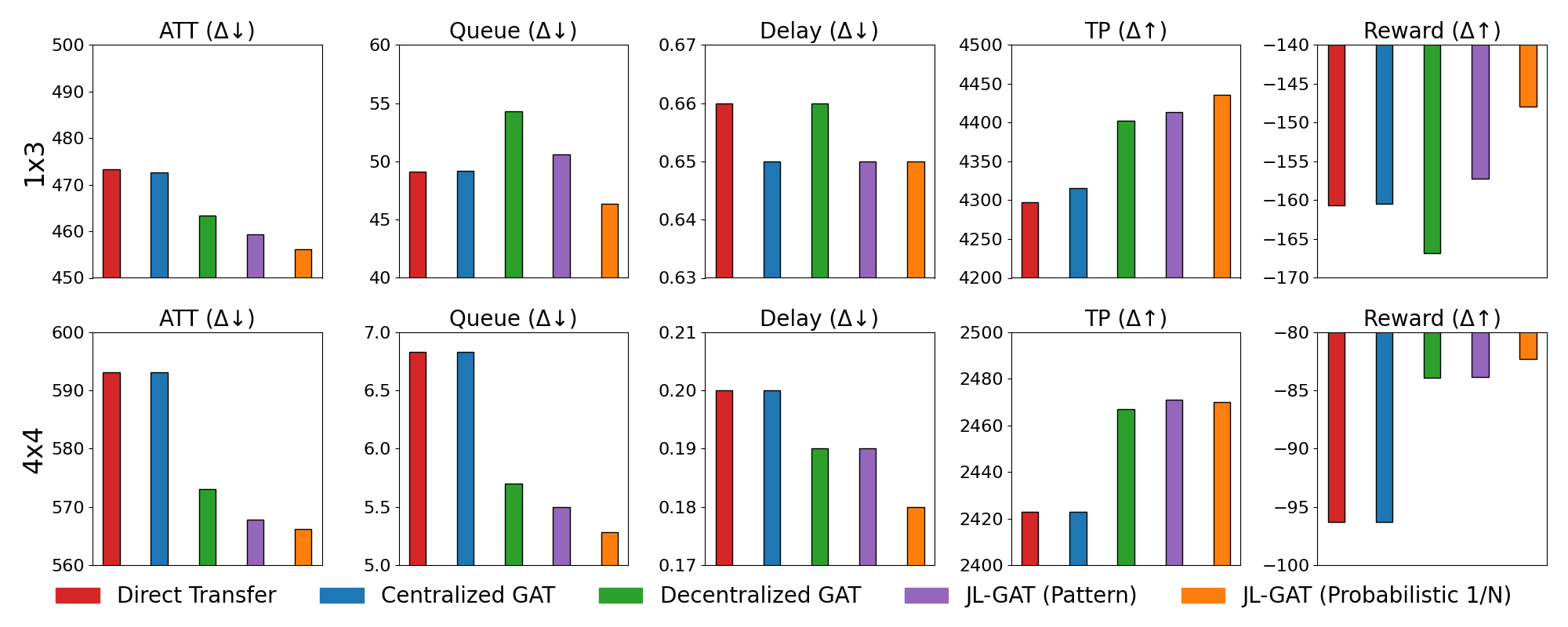}
    \caption{Average performance metrics over the best episode from each trial in the \textbf{snowy} environment. Top row: 1×3 traffic network. Bottom row: 4×4 traffic network. The $\uparrow$ indicates that a higher value represents a better performance for a metric and the $\downarrow$ indicates that a lower value represents a better performance for a metric. Each plot displays the methods in the order they appear from left to right, as indicated in the legend. Full quantitative results, including standard deviations and sim-to-real gap values, are presented in Table~\ref{tab:snowyTable}.}
    \label{fig:mainResults2}
\end{figure}

\section{Dec-POMDP for MARL-based Traffic Signal Control}
\label{sup:decPOMDP}
The traffic signal control (TSC) problem is modeled as a multi-agent reinforcement learning (MARL) task, where each traffic signal operates as an independent agent in a shared environment. The MARL problem is typically formulated as a Decentralized Partially Observable Markov Decision Process (Dec-POMDP), defined by the tuple $\langle \mathcal{N}, \mathcal{S}, \{\mathcal{A}_i\}_{i \in \mathcal{N}}, P, R, \Omega_i, O, \gamma \rangle$, where: $\mathcal{N}$ is the set of agents (intersections), $\mathcal{S}$ is the global state space, representing traffic conditions (e.g., vehicle queues, speeds). $\mathcal{A}_i$ is the action space for agent $i$, which includes actions such as switching traffic signal phases. $P: \mathcal{S} \times \mathcal{A} \to \Delta(\mathcal{S})$ is the transition function, where $\mathcal{A} = \prod_{i \in \mathcal{N}} \mathcal{A}_i$ is the joint action space, and $\Delta(\mathcal{S})$ denotes the set of probability distributions over $\mathcal{S}$. $R: \mathcal{S} \times \mathcal{A} \to \mathbb{R}$ is the reward function, which evaluates traffic metrics (e.g., queue length, delay). $\Omega_i$ is the observation space for agent $i$, with $\Omega = \prod_{i \in \mathcal{N}} \Omega_i$ being the joint observation space. $O$ is the observation probability function $O(s', a, o) = P(o \mid s', a)$ and defines the probability of receiving a joint observation $o$ given then next state $s'$ and joint action $a$. $\gamma \in [0, 1)$ is the discount factor.

At each time step $t$, agent $i$ observes its own state $o_{i,t} \in \Omega_i$, selects an action $a_{i,t} \in \mathcal{A}_i$, and receives a reward $r_{i,t}$. Agent actions are taken simultaneously and comprise a global action $a_{t}$, which transitions the environment from a global state $s_{t}$ to a global next state $s_{t+1}$, where global states consist of observations $o_{i,t}$ for each agent $i$. Global states and actions are represented as:
$s_{t} = (o_1, o_2, \dots, o_N)$, and  $a_{t} = (a_1, a_2, \dots, a_N)$. During training, each agent learns a policy $\pi_i: \Omega_i \to \mathcal{A}_i$ with the goal of maximizing its expected cumulative reward: 
$J_i = \mathbb{E}\left[ \sum_{t=0}^{\infty} \gamma^t r_{i,t} \right]$.

\section{Agent Design Details}
\label{sec:agentDesignDetails}
\begin{itemize}
    \item \textbf{State}. Our state is defined for each agent (intersection) as their own observation $o_{i,t}$ in MARL. For this work, we utilize the state definition from PressLight, simplifying it to include only the number of vehicles in each incoming and outgoing lane without lane segmentation.
    \item \textbf{Action}. Each agent selects an action $a_{i,t} \in \mathcal{A}_i$ at time step $t$ that represents the traffic signal phase $p$. In this work, we utilize the same eight phase TSC action space as in~\citep{da2023uncertainty}, and represent all actions as one-hot encoded vectors.
    \item \textbf{Reward}. The reward $r_{i,t}$ for each agent $i$ at time step $t$ is defined as negative pressure in PressLight. The goal of each agent is to minimize pressure, which effectively balances the number of vehicles in the traffic network and keeps traffic flowing efficiently.
    \item \textbf{Learning Method}. Each agent is trained using an independent Deep Q-Network (DQN) with experience replay, enabling efficient sampling of past experiences. This approach follows established methods in traffic signal control~\citep{wei2018intellilight}. The objective is to optimize the policy $\pi_{i,t}$ for each agent $i$ by using its individual reward $r_{i,t}$ to improve decision-making over time.
\end{itemize}

\section{Code Availability}
The code used in our experiments is publicly available at \href{https://github.com/DaRL-LibSignal/JL-GAT/}{https://github.com/DaRL-LibSignal/JL-GAT/}.

\section{Algorithm for \ours}
\begin{algorithm}
\caption{Algorithm for \ours}\label{alg:jlgatAlgo}

\begin{algorithmic}[0]
\State \textbf{Input:} Initial policies $\pi_{i,\theta}$ for each agent $i$, forward models $f_{i,\phi^{+}}$ for each agent $i$, inverse models $h_{i,\phi^{-}}$ for each agent $i$, simulation dataset $\mathcal{D}_{\text{sim}}$, real-world dataset $\mathcal{D}_{\text{sim}}$, sensing radius $r$, grounding pattern or grounding probability $P_{\text{ground}}^i(t)$ for each agent
\State \textbf{Output:} Policies $\pi_{i,\theta}$, forward models $f_{i,\phi^{+}}$, inverse models $h_{i,\phi^{-}}$
\end{algorithmic}

\begin{algorithmic}[1]
\State Pre-train policies $\pi_{i,\theta}$ for each agent $i$ for $M$ iterations in \(E_{\text{sim}}\)
\For{$e = 1, 2, ..., \textit{I}$}
    \State Rollout policy $\pi_{i,\theta}$ for each agent $i$ in \(E_{\text{sim}}\) and add data to $\mathcal{D}_{\text{sim}}$ (optional)
    \State Rollout policy $\pi_{i,\theta}$ for each agent $i$ in \(E_{\text{real}}\) and add data to $\mathcal{D}_{\text{real}}$ (optional)
    \State \# Update transformation functions for each agent
    \For{$i = 1, 2, ..., N$}
        \State Update $f_{i,\phi^{+}}$ with data from $\mathcal{D}_{\text{real}}$ corresponding to agent $i$ using Equation (\ref{eq:jlgatForwardLoss})
        \State Update $h_{i,\phi^{-}}$ with data from $\mathcal{D}_{\text{sim}}$ corresponding to agent $i$ using Equation (\ref{eq:jlgatInverseLoss})
    \EndFor
    \State \# Policy training
    \For{$ep = 1, 2, ..., \textit{E}$}
        \State \# Action grounding step for each agent $i$ at every time step $t$
        \For{$t = 0, 1, ..., \textit{T-1}$}
            \For{$i = 1, 2, ..., N$}
                \State $a_{i,t}$ = $\pi_{i,\theta}(o_{i,t})$
                \State Predict next state $\hat{o}_{i,t+1}$ using Equation (\ref{eq:jlgatForward})
                \State Calculate grounded action $\hat{{a}}_{i,t}^{\text{g}}$ using Equation (\ref{eq:jlgatInverse})
                \State \# Apply pattern or probabilistic grounding
                \If{grounding is based on a pattern}
                    \State Ground based on a pattern, example shown in Figure~\ref{fig:mainmethod}.
                \ElsIf{grounding is probabilistic}
                    \State Ground with a probability using Equation in Probabilistic Grounding.
                \EndIf
            \EndFor
        \EndFor
        \State \# Policy update step
        \State Improve policies $\pi_{i,\theta}$ for each agent $i$ with reinforcement learning
    \EndFor
\EndFor
\end{algorithmic}
\end{algorithm}

\end{document}